\documentclass[runningheads]{llncs}
\usepackage{graphicx}
\usepackage{amsmath,amssymb} 
\usepackage{color}

\usepackage{subfigure} 
\usepackage{booktabs, array}
\usepackage{epsfig}
\usepackage{multirow}
\usepackage[table,xcdraw]{xcolor}
\usepackage{adjustbox}
\usepackage{hhline}
\usepackage{enumitem}

\begin{document}

\title{Multi-Level Sequence GAN for Group Activity Recognition } 
\titlerunning{MLS-GAN} 


\author{Harshala Gammulle\inst{1} \and
Simon Denman\inst{1}\and
Sridha Sridharan\inst{1} \and Clinton Fookes \inst{1}}
%

\authorrunning{H. Gammulle et al.} 


\institute{ Image and Video Research Laboratory, SAIVT, Queensland University of Technology (QUT), Australia. \\
\email{pranaliharshala.gammulle@hdr.qut.edu.au, \{s.denman, s.sridharan, c.fookes\}@qut.edu.au}}

\maketitle

\begin{abstract}
We propose a novel semi supervised, Multi Level Sequential Generative Adversarial Network (\textit{MLS-GAN}) architecture for group activity recognition. In contrast to previous works which utilise manually annotated individual human action predictions, we allow the models to learn it's own internal representations to discover pertinent sub-activities that aid the final group activity recognition task. The generator is fed with person-level and scene-level features that are mapped temporally through LSTM networks. Action-based feature fusion is performed through novel gated fusion units that are able to consider long-term dependancies, exploring the relationships among all individual actions, to learn an intermediate representation or `action code' for the current group activity. The network achieves it's semi-supervised behaviour by allowing it to perform group action classification together with the adversarial real/fake validation. We perform extensive evaluations on different architectural variants to demonstrate the importance of the proposed architecture. Furthermore, we show that utilising both person-level and scene-level features facilitates the group activity prediction better than using only person-level features. Our proposed architecture outperforms current state-of-the-art results for sports and pedestrian based classification tasks on Volleyball and Collective Activity datasets, showing it's flexible nature for effective learning of group activities. \footnote{This research was supported by the Australian Research Council's Linkage Project LP140100282 ``Improving Productivity and Efficiency of Australian Airports''}
\keywords{Group Activity Recognition  \and Generative Adversarial Networks \and Long Short Term Memory Networks.}
\end{abstract}
\section{Introduction}


	The area of human activity analysis has been an active field within the research community as it can aid in numerous important real world tasks such as video surveillance, video search and retrieval, sports video analytics, etc. In such scenarios, methods with the capability to handle multi-person actions and determine the collective action being performed play a major role. Among the main challenges, handling different personnel appearing at different times and capturing their contribution towards the overall group activity is crucial. Learning the interactions between these individuals further aids the recognition of the collaborative action. Methods should retain the ability to capture information from the overall frame together with information from individual agents. We argue that the overall frame is important as it provides information regrading the varying background and context, the positions of agents within the frame and objects related to the action (e.g. the ball and the net in volleyball) together with the individual agent information.
	
	Recent works on group activity analysis have utilised recurrent neural network architectures to capture temporal dynamics in video sequences. Even though deep networks are capable of performing automatic feature learning, they require manual human effort to design effective losses. Therefore, GAN based networks have become beneficial in overcoming the limitation of deep networks as they are capable of learning both features and the loss function automatically. Furthermore extending the GAN based architecture to a semi-supervised architecture, which is obtained by combining the unsupervised GAN objective with supervised classification objective, leverages the capacity for the network to learn from both labelled and unlabelled data.
	
\begin{figure}[t!]
    \centering
   	 \subfigure[Generator (G)]{\includegraphics[width=.5\linewidth]{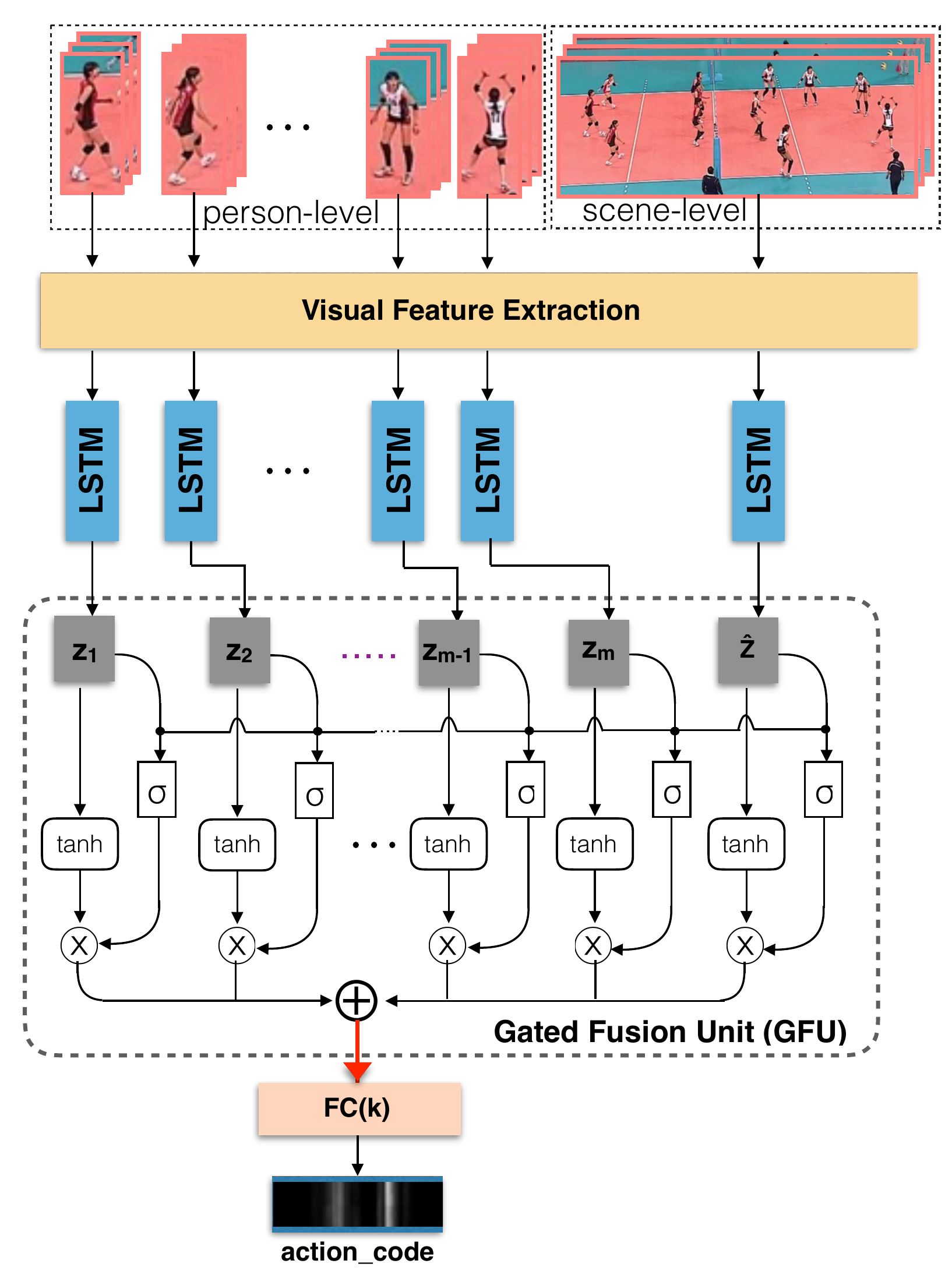}}
    	  \subfigure[Discriminator (D)]{\includegraphics[width=.35\linewidth]{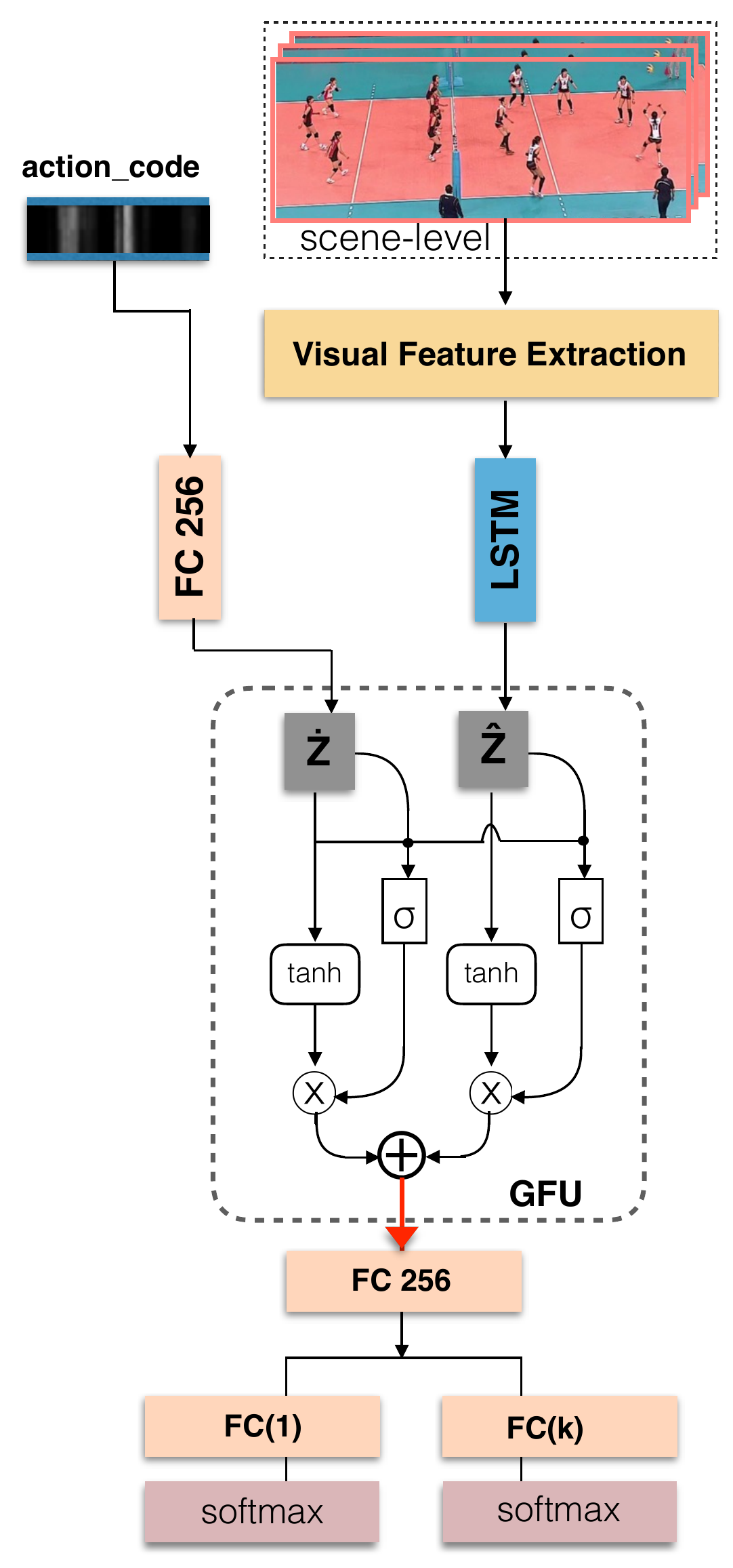}}
    \caption{The proposed Multi-Level Sequence GAN (\textit{MLS-GAN}) architecture: (a) G is trained with sequences of person-level and scene-level features to learn an intermediate action representation, an `action code'. (b) The model D performs group activity classification while discriminating real/fake data from scene level sequences and ground truth/generated action codes.}
    \label{fig:framework}
\end{figure}

	In this paper we present a semi-supervised Generative Adversarial Network (GAN) architecture based on video sequence modelling with LSTM networks to perform group activity recognition. Figure \ref{fig:framework}, shows the overall framework of our proposed GAN architecture for group activity recognition. The generator is fed with sequences of person-level and scene-level RGB features which are extracted through the visual feature extractor for each person and the scene. Then the extracted features are sent through separate LSTMs to map the temporal correspondence of the sequences at each level. We utilise a gated fusion unit inspired by \cite{gatedfusion} to map the relevance of these LSTM outputs to an intermediate action representation, an `action code', to represent the current group action. These action codes are then employed by the discriminator model which determines the current group action class and whether the given action code is real (ground truth) or fake (generated). Overall, the generator focuses on generating action codes that are indistinguishable from the ground truth action codes while the discriminator tries to achieve the real/fake and group activity classifications. With the use of a gated fusion unit, the model gains the ability to consider all the inputs when deciding on the output. Therefore, it is able to map the relevance of each performed individual action and their interactions with the attention weights automatically, to perform the final classification task. The contributions of our proposed model are as follows: (i) we introduce a novel recurrent semi-supervised GAN framework for group activity recognition, (ii) we formulate the framework such that it incorporates both person-level and scene-level features to determine the group activity, (iii)  we demonstrate a feature fusion mechanism with gated fusion units, which automatically learns an attention mechanism focusing on the relevance of each individual agent and their interactions, (iv) we evaluate the proposed model on pedestrian and sports datasets and achieve state-of-the-art results, outperforming the current baseline methods. 
	

\section{Related Work}

Human action recognition is an area that has been of pivotal interest to researchers in the computer vision domain. However, a high proportion of proposed models are based on single-human actions which do not align with the nature of the real world scenarios where actions are continuous. Furthermore, many existing approaches only consider actions performed by a single agent, limiting the utility of these approaches. 

Some early works \cite{amer2014,choi2009,lan2012social,ramanathan2013} on group activity recognition have addressed the group activity recognition task on surveillance and sports video datasets with probabilistic and discriminative models that utilise hand-crafted features. As these hand-crafted feature based methods always require feature engineering, attention has shifted towards deep network based methods due to their automatic feature learning capability.

In \cite{ibrahim2016cvpr} authors introduce an LSTM based two stage hierarchical model for group activity recognition. The model first learns the individual actions which are then integrated into a higher level model for group activity recognition. Shu et al. in \cite{CERN_cvpr} have introduced the Confidence Energy Recurrent Network (CERN) which is also a two-level hierarchy of LSTM networks that utilises an energy layer for estimating the energy of the predictions. As these LSTM based methods focus on learning each individual action independently, afterwhich the group activity is learnt by considering the predicted individual action, they are unable to map the interactions between individuals well \cite{SRNN_wacv2018}. Kim et al. \cite{kim2018} proposed a gated recurrent unit (GRU) based model that utilises discriminative group context features (DGCF) to handle people as individuals or sub groups. Another similar approach is suggested in \cite{puck_possession} for classifying puck possession events in ice hockey by extracting convolutional layer features to train recurrent networks. In \cite{SRNN_wacv2018}, the authors introduced the Structural Recurrent Neural Network (SRNN) model which is able to handle a varying number of individuals in the scene at each time step with the aid of a grid pooling layer. Even though these deep network based models are capable of performing automatic feature learning, they still require manual human effort to design effective losses.

Motivated by the recent advancements and with the ability to learn effective losses automatically, we build on the concept of Generative Adversarial Networks (GANs) to propose a recurrent semi-supervised GAN framework for group activity recognition. GAN based models are capable of learning an output that is difficult to discriminate from real examples, and also learn a mapping from input to output while learning a loss function to train the mapping. As a result of this ability, GANs have been used in solving different computer vision problems such as inpainting \cite{inpainting}, product photo generation \cite{yoo2016} etc. We utilise an extended variant of GANs, the conditional GAN \cite{condGAN,fernando2018tracking,fernando2018task} architecture where both the generator and the discriminator models are conditioned with additional data such as class labels or data from other modalities. A further enhancement of the architecture can be achieved by following the semi-supervised GAN architecture introduced in \cite{denton2016semi}. There are only a handful of GAN based methods \cite{GAN_action_2,GAN_action_1} that have been introduced for human activity analysis. In \cite{GAN_action_2} the authors train the generative model to synthesise frames in an action video sequence, and in \cite{GAN_action_1} the generative model synthesises masks for humans. While these methods try to learn a distribution on video frame level attributes, no effort has been made to learn an intermediate representation at the human behaviour level. Motivated by \cite{infogail_nips,bora2017}, which have demonstrated the viability of learning intermediate representations with GANs, we believe that learning such an intermediate representation (`action code') would help the action classification process, as the classification model has to classify this discriminative action code.

To this end we make the first attempt to apply GAN based methods to the group activity recognition task, where the network jointly learns a loss function as well as providing auxiliary classification of the class. 
\section{Methodology}

GANs are generative models that are capable of learning a mapping from a random noise vector $z$ to an output vector, $y: G:z\rightarrow y$ \cite{GAN2014}. In our work, we utilise the conditional GAN \cite{condGAN}, an extension of the GAN that is capable of learning a mapping from the observed image $x_{t}$ at time $t$ and a random noise vector $z_{t}$ to $y_{t}: G:\{x_{t},z_{t}\}\rightarrow y_{t}$ \cite{condGAN,Isola_CVPR2017}. 

GANs are composed of two main components: the Generator (G) and the Discriminator (D), which compete in a two player game. G tries to generate data that is indistinguishable from real data, while D tries to distinguish between real and generated (fake) data.    

We introduce a conditional GAN based model, Multi-Level Sequence GAN (\textit{MLS-GAN}), for group activity recognition, which utilises sequences of person-level and scene-level data for classification. In Section \ref{sec:ac_code}, we describe the action code format that the GAN is trained to generate; Section \ref{sec:semi_gan} describes the semi-supervised GAN architecture; and in Section \ref{sec:objective} we explain the objectives that the models seek to optimise. 

\subsection{Action codes}
\label{sec:ac_code}

\begin{figure}[t!]
    \centering
    \subfigure[] {\includegraphics[width=.44\linewidth]{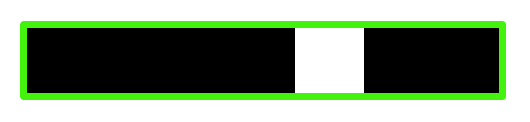}}
     \subfigure[] {\includegraphics[width=.45\linewidth]{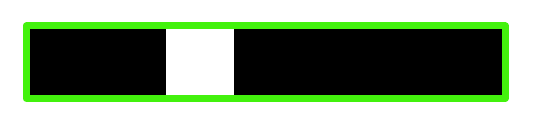}}
    \caption{Sample ground truth action codes, with $k=7$ (i.e. we have 7 actions). For the code in (a), $y=5$ and for the code shown in (b) $y=3$. Note that a green border is shown around the codes for clarify, this is not part of the code and is only included to aid display. Codes are of size 1 x k pixels.}
    \label{fig:action_codes}
\end{figure}

The generator network is trained to synthesise an `action code' to represent the current group action. The generator maps dense pixel information to this action code. Hence having a one hot vector is not optimal. Therefore we scale it to a range from 0 to 255 giving more freedom for the action generator and discriminator to represent each action code as a dense vector representation, 
\begin{equation}
y  \epsilon  {\rm I\!R}^{1 \times{k}},
\label{eq:1}
\end{equation} 

where k is the number of group action classes in the dataset. This action code representation can also be seen as an intermediate representation for the action classification. Several works have previously demonstrated the viability of these representations with GANs \cite{infogail_nips,bora2017}. Overall, this action code generation is effected by adversarial loss as well as the classification loss, where the learnt action codes need to be informative for the classification task. In Figure \ref{fig:action_codes} we have sample action codes for a scenario where there are 7 action classes.

\subsection{Semi-supervised GAN architecture}
\label{sec:semi_gan}

The semi-supervised GAN architecture is achieved by combining the unsupervised GAN objective with the supervised classification objective. Unlike the standard GAN architecture, the discriminator of the semi-supervised GAN network is able to perform group action classification together with the real/fake classification task.

\subsubsection{Generator}

The generator takes person-level inputs and scene-level inputs for each video sequence for T time steps. Let $\{\hat{X_1},\hat{X_2},\ldots,\hat{X_T}\}$ be the full frame image sequences, while $\{ x^n_1, x^n_2,\ldots,x^n_T \}$ is the person-level cropped bounding box image sequence for the $n^{th}$ person, where $n \in [1,\ldots,N]$. The generator input, $I_{G}$, can be defined as follows,

\begin{equation}
I_{G}=(\{ x^1_1, x^1_2,\ldots,x^1_T \},\ldots,\{ x^N_1, x^N_2,\ldots,x^N_T \},\{\hat{X_1},\hat{X_2},\ldots,\hat{X_T}\}).
\label{eq:1}
\end{equation}   

The generator learns a mapping from the observed input $I_{G}$ and a noise vector $z$ to $y : G : \{I_{G},z\}\rightarrow y$, where $y$ is the action code. As shown in Figure \ref{fig:framework}, to obtain this mapping the generator extracts visual features from $I_{G}$ using a pre-trained deep network. Let the extracted scene-level visual feature be $\hat{\theta}_{t}$ and the extracted person-level features be $\theta^n_{t}$ for the $n^{th}$ person such that,

\begin{equation}
\hat{\theta}_{t}=f(\hat{X_t}),
\label{eq:2}
\end{equation}   
and
\begin{equation}
\theta^n_{t}=f(x^n_t).
\label{eq:3}
\end{equation}  

These extracted visual features are then passed through the respective LSTM networks,

\begin{equation}
\hat{Z}=\mathrm{LSTM}([\hat{\theta}_{1},\hat{\theta}_{2},\ldots,\hat{\theta}_{T}]),
\label{eq:4}
\end{equation}  

\begin{equation}
Z^n=\mathrm{LSTM}([\theta^n_{1},\theta^n_{2},\ldots,\theta^n_{T}]).
\label{eq:5}
\end{equation}  

Outputs for the $n^{th}$ person LSTM model are subsequently sent through a gated fusion unit to perform feature fusion as follows,

\begin{equation}
h^n=\tanh(\dot{W}^n Z^n),
\label{eq:6}
\end{equation}  

where $\dot{W}^n$ is a weight vector for encoding. Next the sigmoid function, $\sigma$, is used to determine the information flow from each input stream,  

\begin{equation}
q^n=\sigma(\bar{W}^n[Z^1, Z^2,\ldots, Z^N, \hat{Z}]),
\label{eq:7}
\end{equation}  
afterwhich we multiply the embedding $h^{n}$ with gate output $q^n$ such that,
\begin{equation}
r^n=h^n \times{q^n}.
\label{eq:8}
\end{equation}  

Therefore, when determining information flow from the $n^{th}$ person  stream we attend over all the other input streams, rather than having one constant weight value for the entire stream. Using these functions we generate gated outputs, $r^n$, for each person level input as well as the other $\hat{r}$ for the scene level input. Given these person and scene level outputs, the fused output of the gated unit can be defined as,

\begin{equation}
C=\sum_{j=1}^{N} r^n + \hat{r}.
\label{eq:9}
\end{equation}  

This output, $C$, is finally sent through a fully connected layer to obtain the action code to represent the current action,

\begin{equation}
y=FC(C, z),
\label{eq:10}
\end{equation}  
which also utilises a latent vector $z$ in the process.
\subsubsection{Discriminator}

The discriminator takes the scene-level inputs ($I_{D1}$) from each video sequence for T time steps together with real (ground truth)/ fake (generated) action codes ($I_{D2}$). The aim of the semi-supervised GAN architecture is to perform real/fake validation together with the group action classification. The inputs to the discriminator models are as follows,

\begin{equation}
I_{D1}=(\{\hat{X_1},\hat{X_2},\ldots,\hat{X_T}\}) ; I_{D2}=y.
\label{eq:11}
\end{equation}   

Unlike the generator, the discriminator is not fed with person-level features. The action codes provide intermediate representations of the group activities that have been generated by considering person-level features. Therefore, the activities of the individuals are already encoded in the action codes and the scene-level features are used to support the decision. Considering these scene level inputs also contain the individual people, providing the crops of every individual is redundant and greatly increases the architecture complexity. We believe that by providing the scene level features the model should be able to capture the spatial relationships and the spatial arrangements of the individuals, which is essential when deciding upon the authenticity of the generated action code.  

The scene-level feature input ($I_{D1}$) is then sent through the visual feature extractor defined in Equation \ref{eq:2} and we obtain $\hat{\theta}_{t}$. The scene-level features capture spatial relationships and the spatial arrangements of the people, which helps to decide whether the action is realistic given the arrangements. The action code input ($I_{D2}$) is sent through a fully connected layer and we obtain $\acute{\theta}_{t}$. These extracted features are then sent through gated fusion unit to perform feature fusion and the output of the gated unit $\acute{C}$ can be defined as,

\begin{equation}
\acute{C}=\acute{r}^n + \hat{r}^n
\label{eq:12}
\end{equation}  

Finally $\acute{C}$ is passed through fully connected layers to perform group action classification together with the real/fake validation of the current action code.
 
 \subsection{GAN Objectives}
 \label{sec:objective}
 
 The objective of the proposed \textit{MLS-GAN} model can be defined as,
 
\begin{equation}
\begin{split}
L_{GAN}(G,D)= \mathop{min}_{G} \mathop{max}_{D} \hspace{1mm}\mathbb{E}[\log D(I_{D1},I_{D2})]+ \\ \mathbb{E}[\log(1-D(I_{D1},G(I_{G},z))]+ \lambda_{c} \mathbb{E}[\log D_{c}(k|I_{D1},I_{D2})],  
\end{split}
\label{eq:13}
\end{equation}    
where $D_c(x)$ is the output classifier head of the discriminator and  $\lambda_{c}$ is a hyper parameter which balances the contributions of classification loss and the adversarial loss.


\section{Experiments}

\subsection{Datasets}

To demonstrate the flexibility of the proposed method we evaluate our proposed model on sports and pedestrian group activity datasets: the volleyball dataset \cite{ibrahim2016cvpr} and the collective activity dataset \cite{choi2009}. We don't use the annotation for individual person activities in this research. Rather, we allow the model to learn it's own internal representation of the individual activities. We argue this is more appropriate for group activity recognition as the model is able to discover pertinent sub-activities rather than being forced to learn a (possibly) less informative representation that is provided by the individual activity ground truth.  


\subsubsection{Volleyball dataset}

The Volleyball dataset is composed of 55 videos containing 4,830 annotated frames. The dataset represents 8 group activities that can be found in Volleyball : right set, right spike, right pass, right win-point, left win-point, left pass, left spike and left set. The train/test splits of \cite{ibrahim2016cvpr,CERN_cvpr} are used.       
   
\subsubsection{Collective activity dataset}

The collective activity dataset is composed of 44 video sequences representing five group-level activities. The group activity label is assigned by considering the most common action that is performed by the people in the scene. The train/test splits for evaluations are as in \cite{ibrahim2016cvpr}. The available group actions are crossing, walking, waiting, talking and queueing.   

\subsection{Metrics}

We perform comparisons to the state-of-the-art by utilising the same metrics used by the baseline approaches \cite{latent_SVM,ibrahim2016cvpr}. We use the multi-class accuracy (MCA) and the mean per class accuracy (MPCA) to overcome the imbalance in the test set (e.g. the total number of crossing examples is more than twice that of queueing and talking examples \cite{latent_SVM}) when evaluating the performance. As MPCA calculates the accuracy for each class, before taking the average accuracy values, this overcomes the accuracy bias on the imbalanced test set. 

\subsection{Network Architecture and Training}

We extract visual features through a ResNet-50 \cite{resnet} network pre-trained on ImageNet \cite{imageNet} for each set of person-level and scene-level inputs. Each input frame is resized to 224 x 224 as a preprocessing step prior to feature extraction. The features are extracted from the $40^{th}$ layer of ResNet-50 and these features are then sent through the first layer of the LSTMs which have 10 time steps. The number of LSTMs for the first layer is determined by considering the maximum number of persons (with bounding boxes) in each dataset. If the maximum number of available bounding boxes for a dataset is N, then the first layer of LSTMs is composed of (N+1) LSTMs i.e. one LSTM for each person plus one for the scene level features. In cases where there are fewer than N person we create a dummy sequence with default values. We select $N=12$ for the volleyball dataset and $N=10$ for the collective activity dataset. The gated fusion mechanism automatically learns to discard dummy sequences when there are less than N people in the scene. 

The outputs of these LSTMs are passed through the gated fusion unit (GFU) to map the correspondences among person-level and scene-level streams. For all the LSTMs we set the hidden state embedding dimension to be 300 units. For the volleyball dataset the dimensionality of the FC(k) layer is set to 8 as there are 8 group activities in the dataset, and for the collective activity dataset we set this to 5. The hyper parameter, $\lambda_c=2.5$, is chosen experimentally. 

In both datasets, the annotations are given in a consistent order. In the volleyball dataset the annotations are ordered based on player role (i.e. spiker, blocker); and in the collective dataset, persons in the frame are annotated from left to right in the scene. We maintain this order of the inputs allowing the GFU to understand the contribution of each person in the scene and learn how the individual actions affect the group action.

The training procedure is similar to \cite{Isola_CVPR2017} and alternates between one gradient decent pass for the discriminators and one for the action generators using mini-batch standard gradient decent (32 examples per mini-batch), and uses the Adam optimiser \cite{adam2015} with an initial learning rate of 0.1 for 250 epochs and 0.01 for the next 750 epochs.

For discriminator training, we take (batch\_size)/2 generated (fake) action codes and (batch\_size)/2 ground truth (real) action codes where the ground truth action codes are manually created. We use Keras \cite{keras} and Theano \cite{theano} to implement our model.

\subsection{Results}

Table \ref{tab:table1} and \ref{tab:table2} present the evaluations for the proposed \textit{MLS-GAN} along with the state-of-the-art baseline methods for the  Collective Activity \cite{choi2009} and Volleyball \cite{ibrahim2016cvpr} benchmark datasets respectively. 

When observing the results in Table \ref{tab:table1}, we observe poor performance from the hand-crafted feature based models \cite{latent_SVM,cardinality_kernel} as they are capable of capturing only abstract level concepts \cite{gammulle2017}. The deep structured model \cite{deng2015} utilising a CNN based feature extraction scheme improves upon the handcrafted features. However it does not utilise temporal modelling to map the evolution of the actions, which we believe causes the deficiencies in it's performance.

The authors in  \cite{ibrahim2016cvpr,CERN_cvpr} utilise enhanced temporal modelling through LSTMs and achieved improved performance. However we believe the two step training process leads to an information loss. First, they train a person-level LSTM model which generates a probability distribution over the individual action class for each person in the scene. In the next level only these distributions are used for deciding upon the group activities. Neither person-level features, nor the scene structure information such as the locations of the individual persons is utilised. 

In contrast, by utilising features from both the person level and scene level, and further improving the learning process through the proposed GAN based learning framework, the proposed MLS-GAN model has been able to outperform the state-of-the-art models in both considered metrics. 

\begin{table}[htbp]
\centering
\caption{Comparison of the results on Collective Activity dataset \cite{choi2009} using MCA and MPCA. NA refers to unavailability of that evaluation.}
\label{tab:table1}
\begin{tabular}{|p{6cm}|c|c|}
\hline
                                                                                   Approach                              & MCA                               & MPCA                              \\ \hline \hline

                                                                                          Latent SVM \cite{latent_SVM}                            & 79.7                              & 78.4                              \\ \hline
                                                                                          Deep structured \cite{deng2015}                       & 80.6                              &    NA                               \\ \hline
                                                                                          Cardinality Kernel \cite{cardinality_kernel}                    & 83.4                              & 81.9                              \\  \hline
                                                                                          2-layer LSTMs \cite{ibrahim2016cvpr}                         & 81.5                              & 80.9                              \\ \hline
                                                                                          CERN \cite{CERN_cvpr}                                  & 87.2                              & 88.3                              \\ \hline \hline
\cellcolor[HTML]{C0C0C0}MLS-GAN & \cellcolor[HTML]{C0C0C0}\textbf{91.7} & \cellcolor[HTML]{C0C0C0}\textbf{91.2} \\ \hline
\end{tabular}
\end{table}

\begin{figure}[htbp]
 \centering
    \subfigure{\includegraphics[width=.23\linewidth,height=2.0cm]{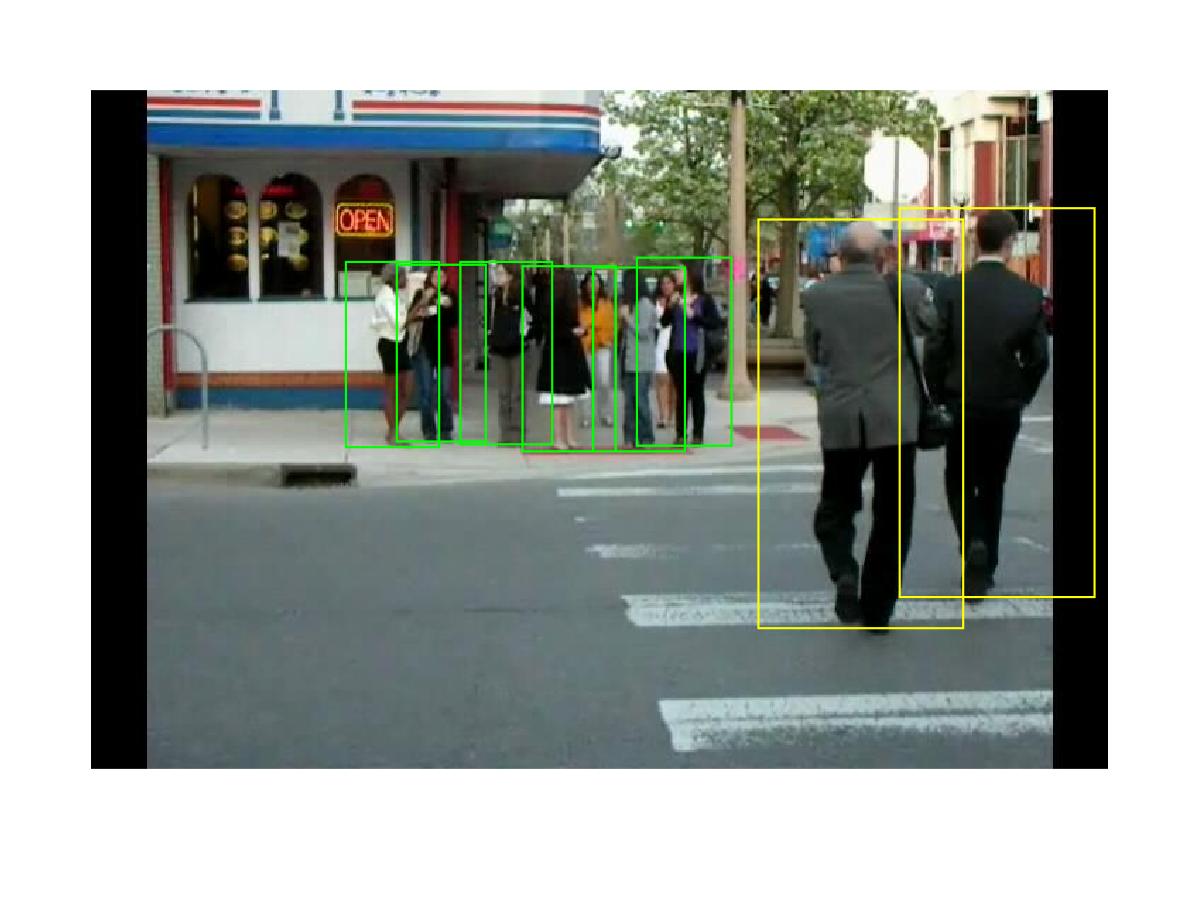}}
    \subfigure{\includegraphics[width=.23\linewidth,height=2.0cm]{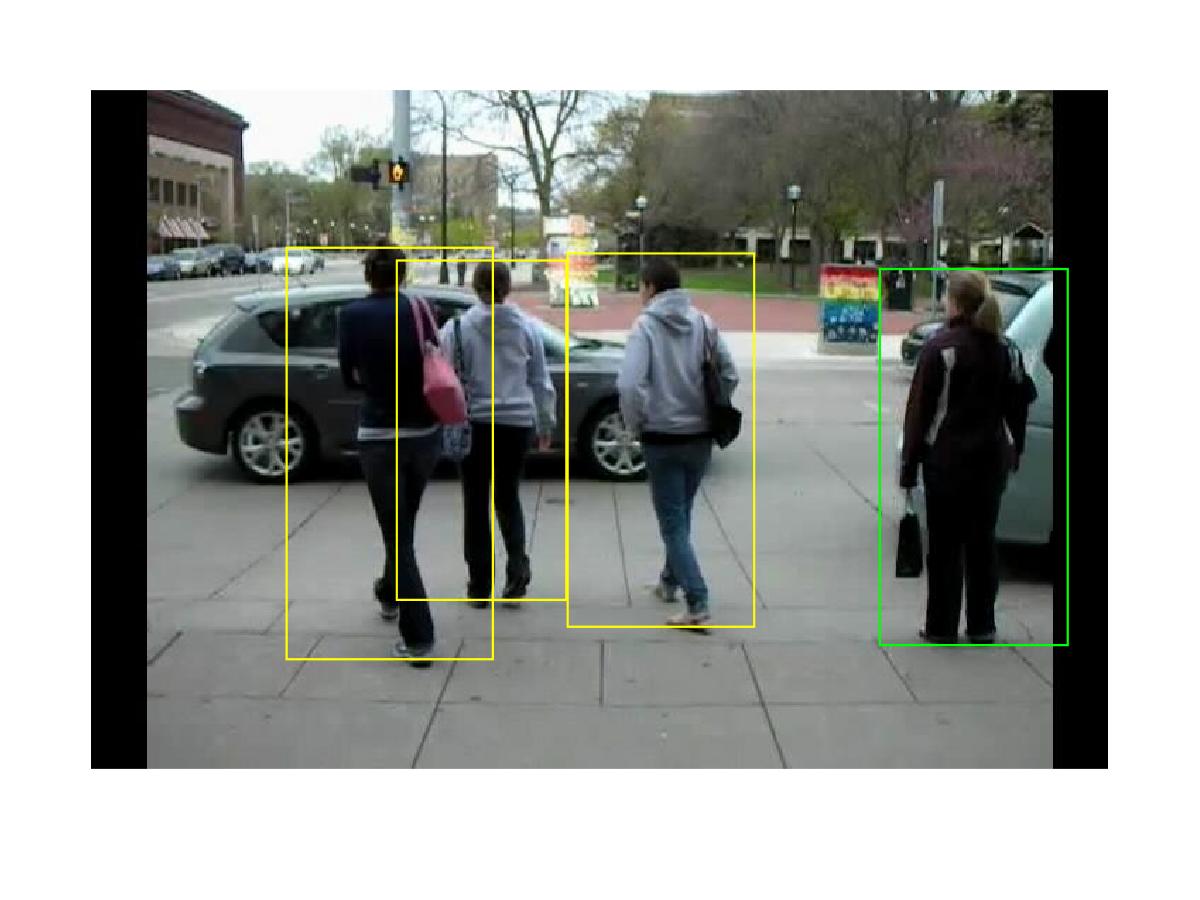}}
    \subfigure{\includegraphics[width=.23\linewidth,height=2.0cm]{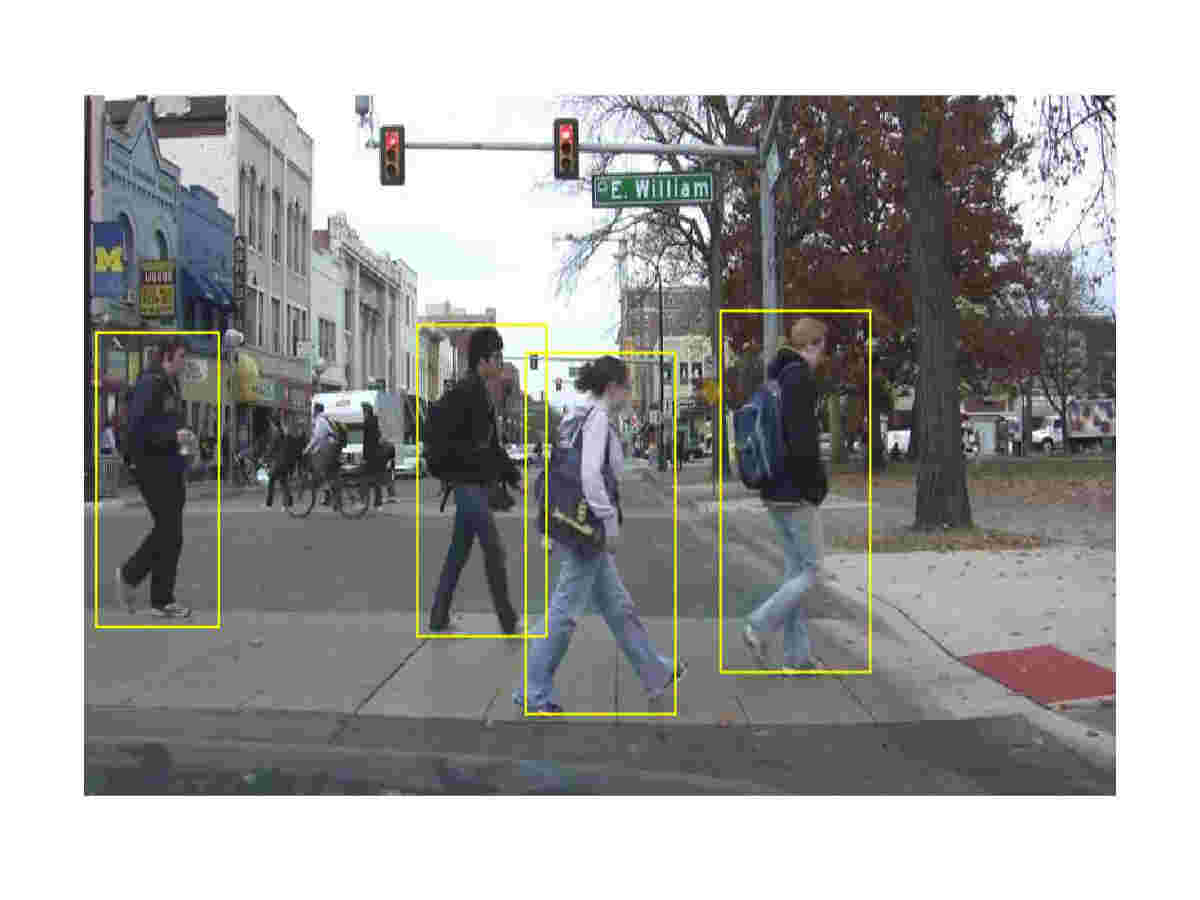}}
    \subfigure{\includegraphics[width=.23\linewidth,height=2.0cm]{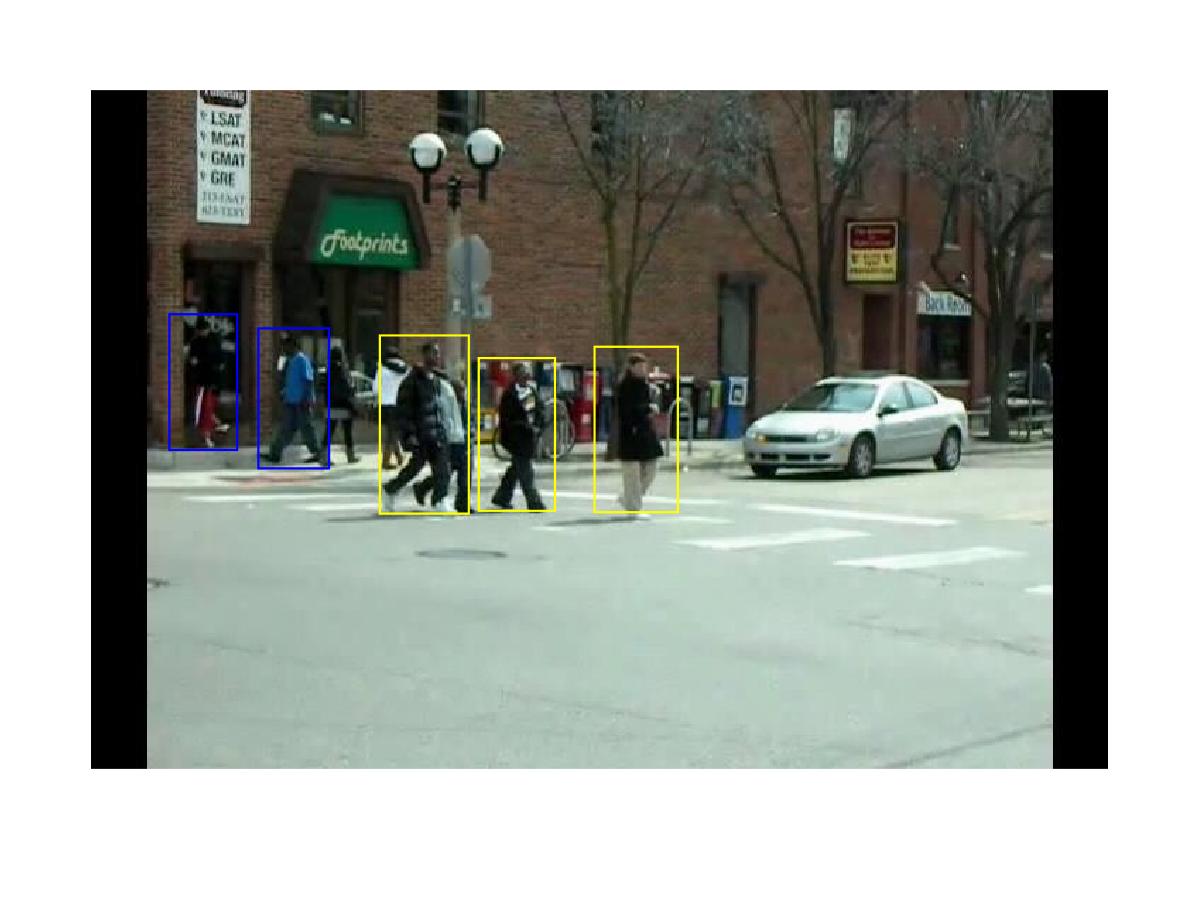}}
     \subfigure{\includegraphics[width=.23\linewidth,height=2.0cm]{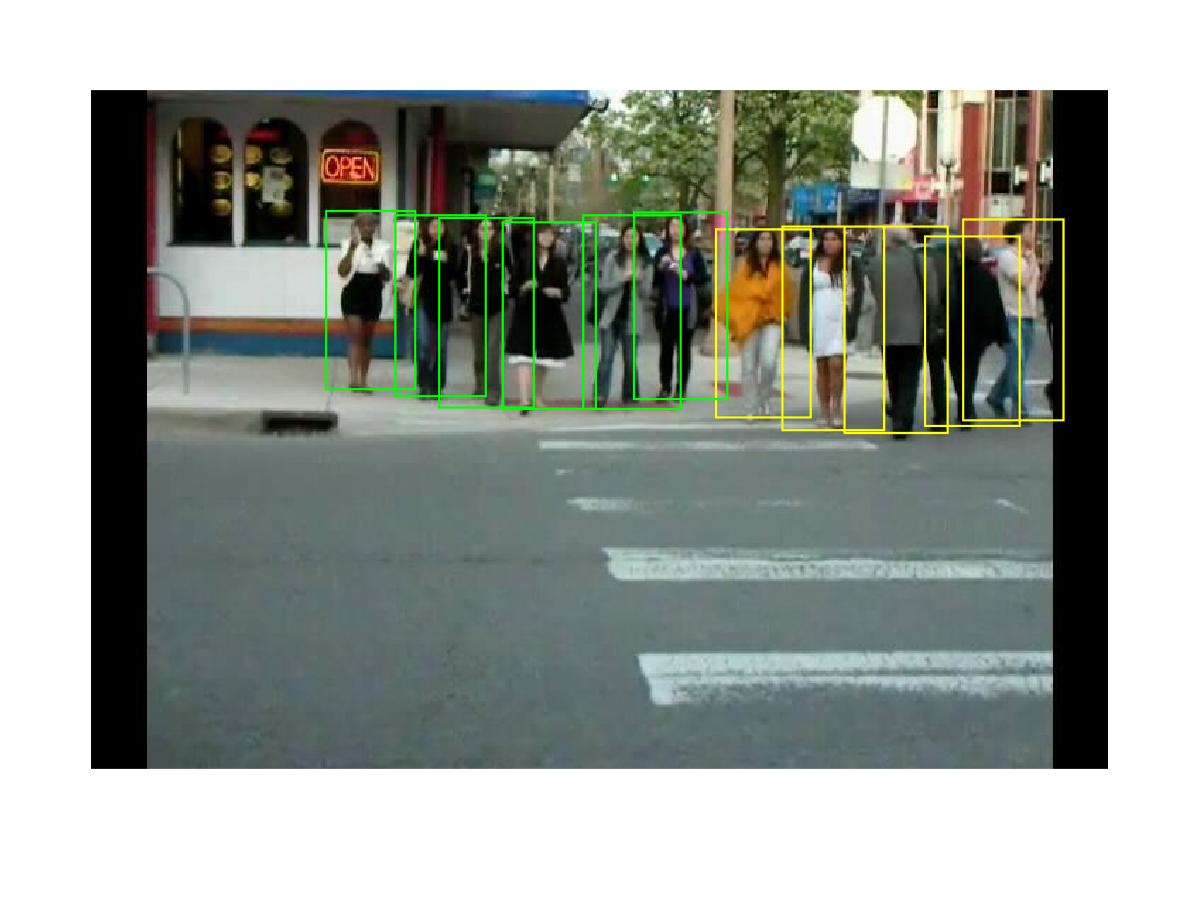}}
    \subfigure{\includegraphics[width=.23\linewidth,height=2.0cm]{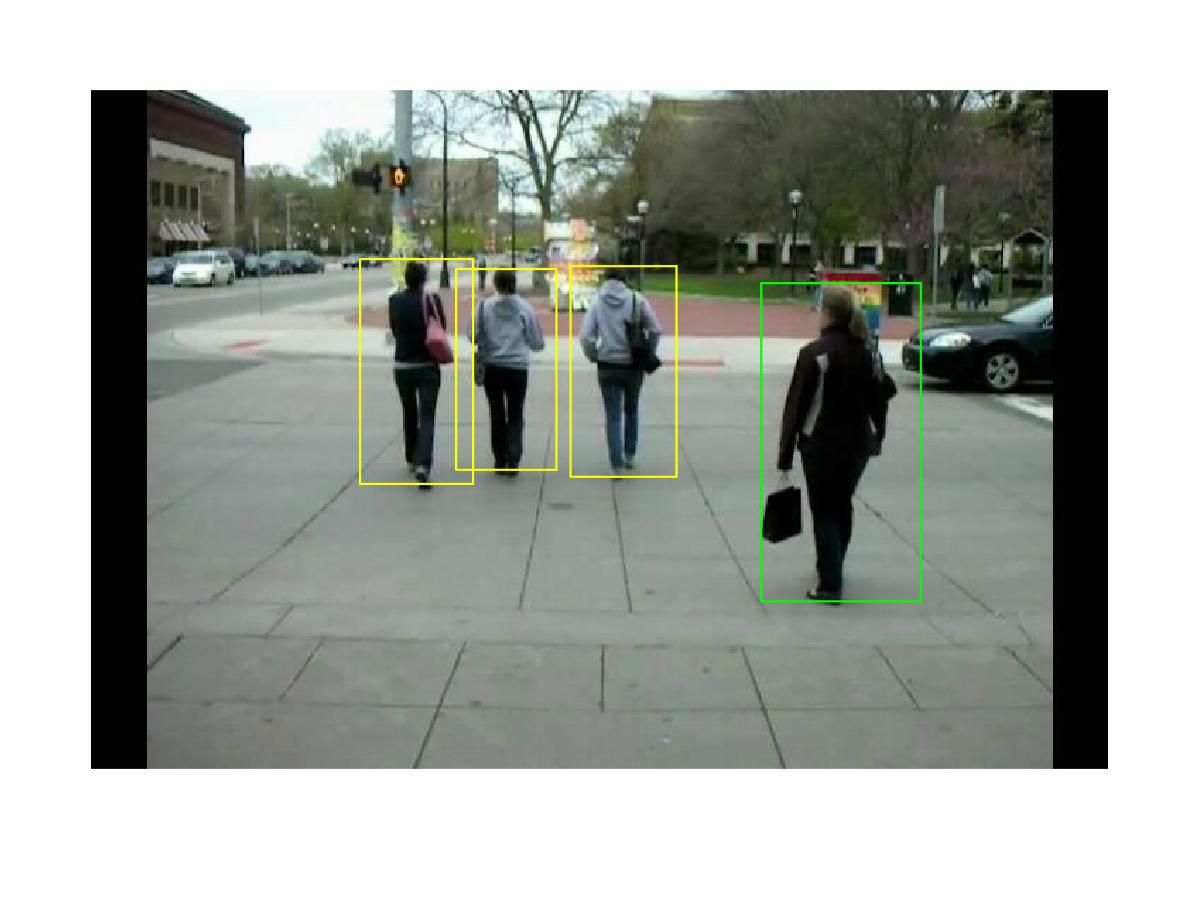}}
    \subfigure{\includegraphics[width=.23\linewidth,height=2.0cm]{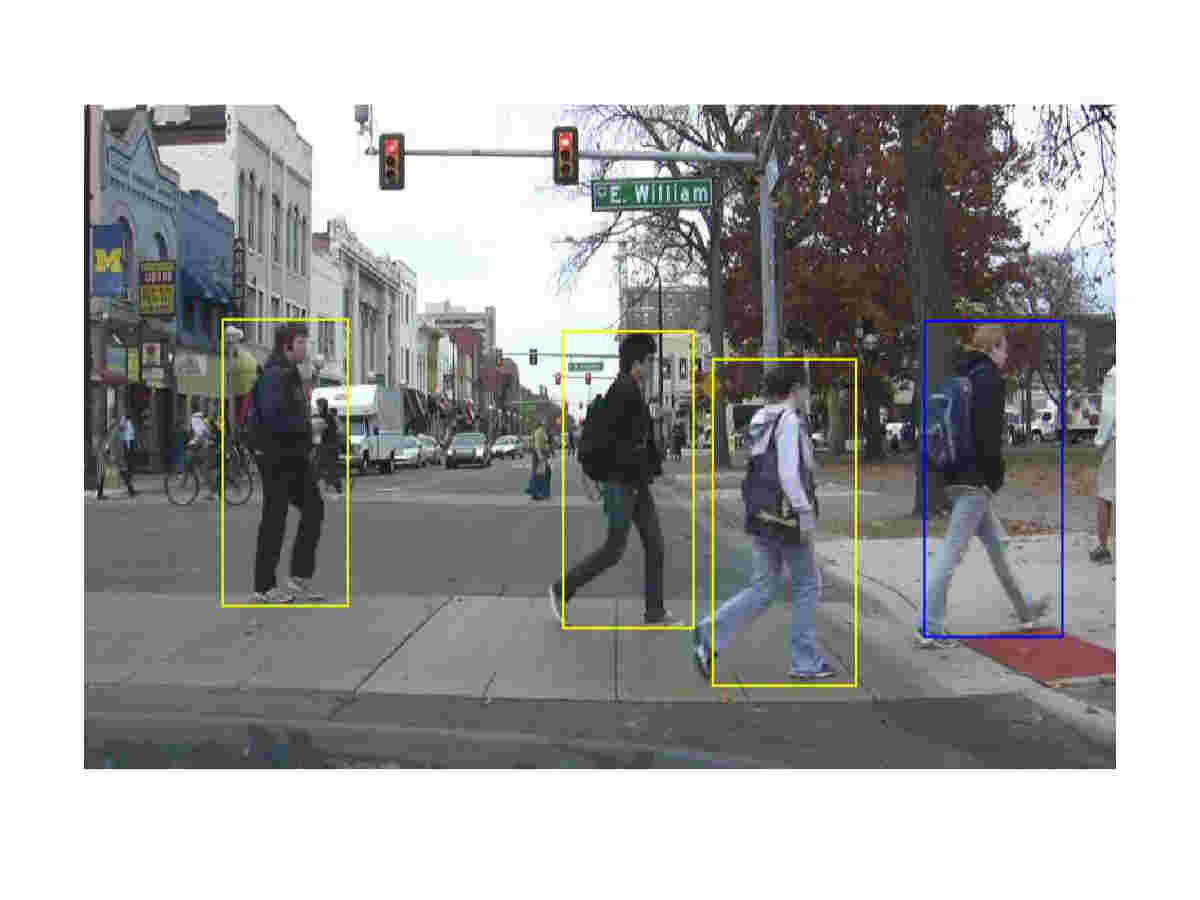}}
    \subfigure{\includegraphics[width=.23\linewidth,height=2.0cm]{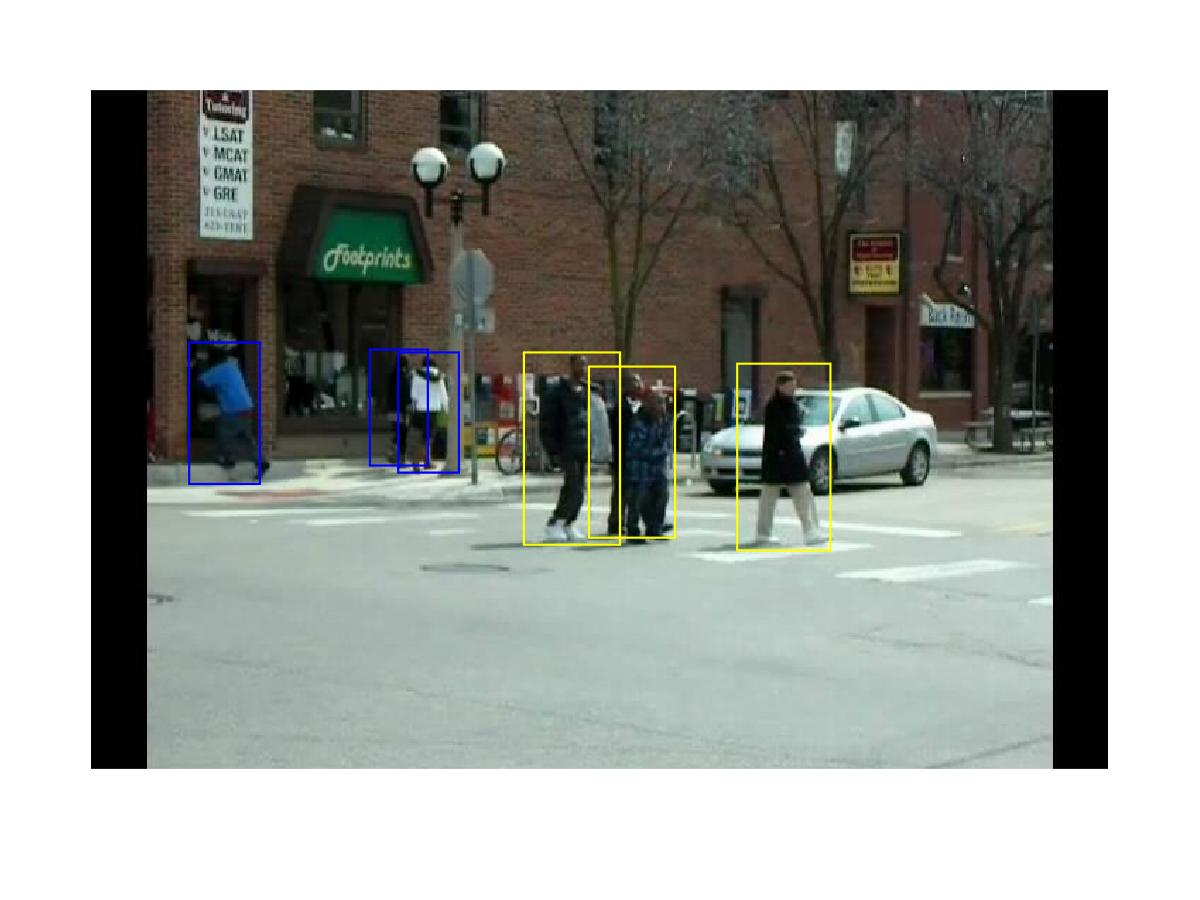}}
    \subfigure{\includegraphics[width=.23\linewidth,height=2.0cm]{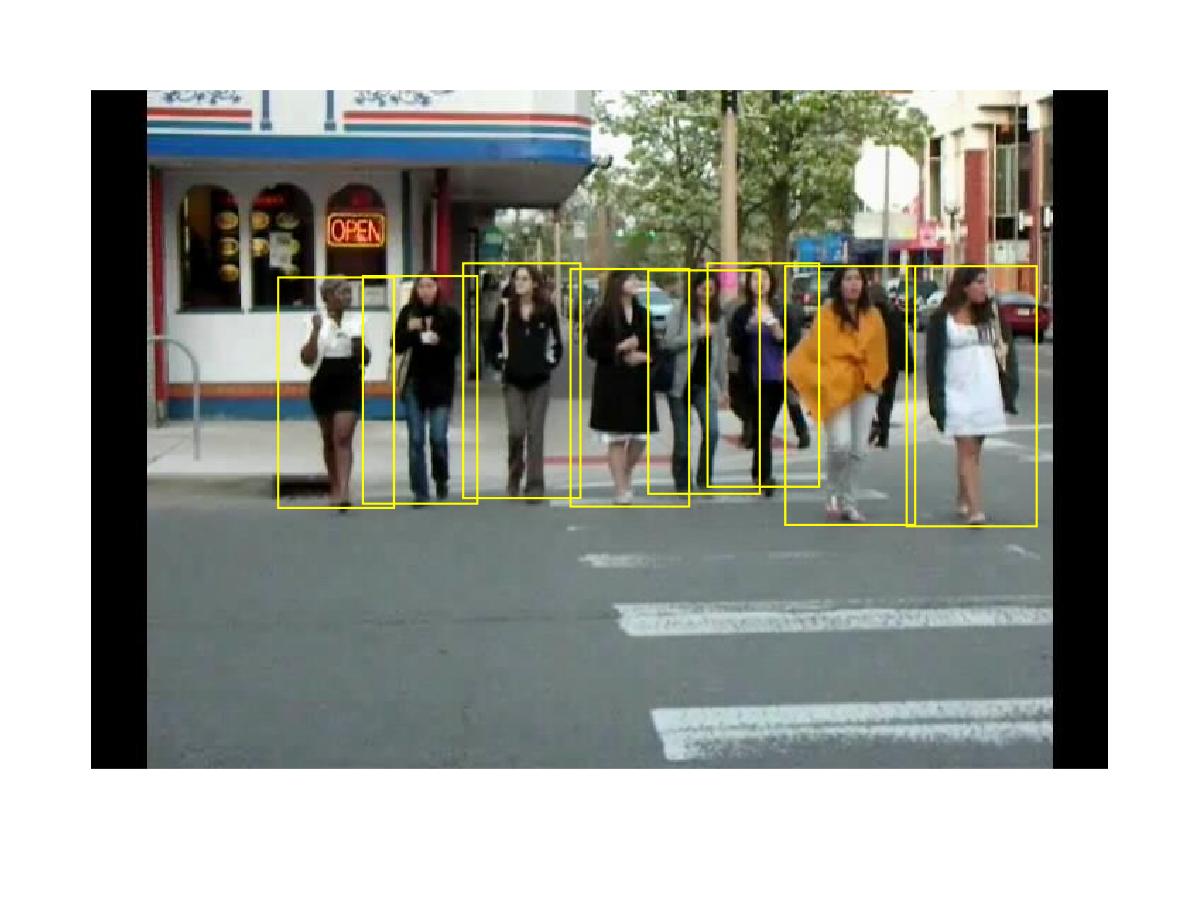}}
    \subfigure{\includegraphics[width=.23\linewidth,height=2.0cm]{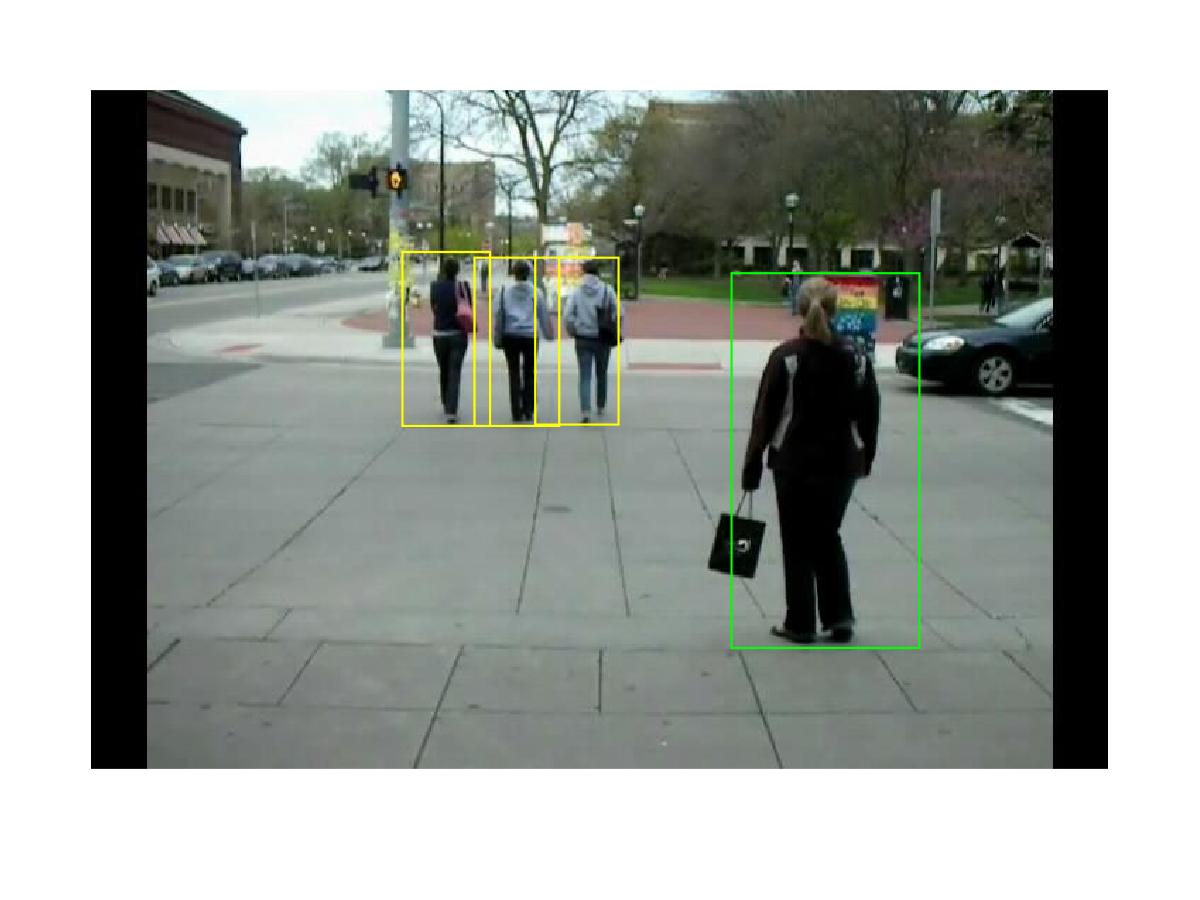}}
    \subfigure{\includegraphics[width=.23\linewidth,height=2.0cm]{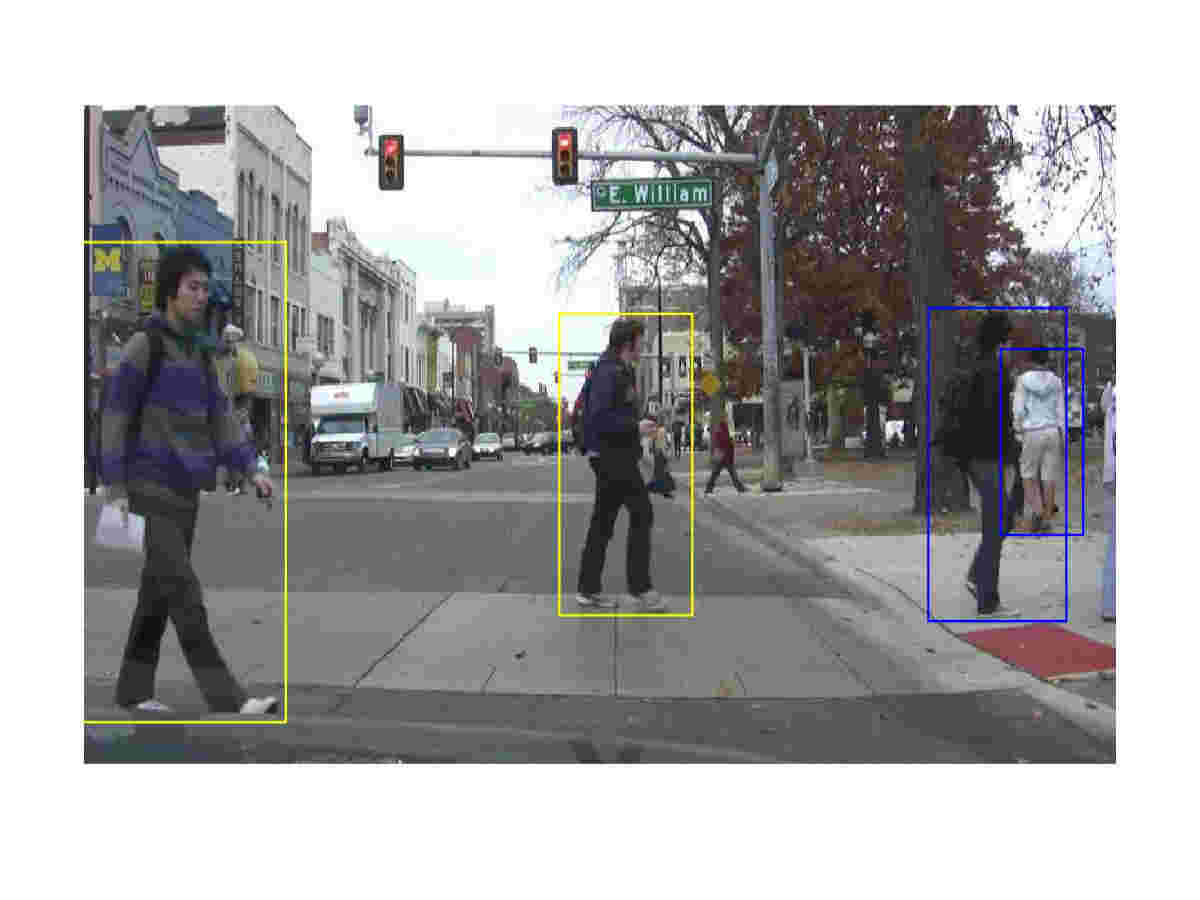}}
    \subfigure{\includegraphics[width=.23\linewidth,height=2.0cm]{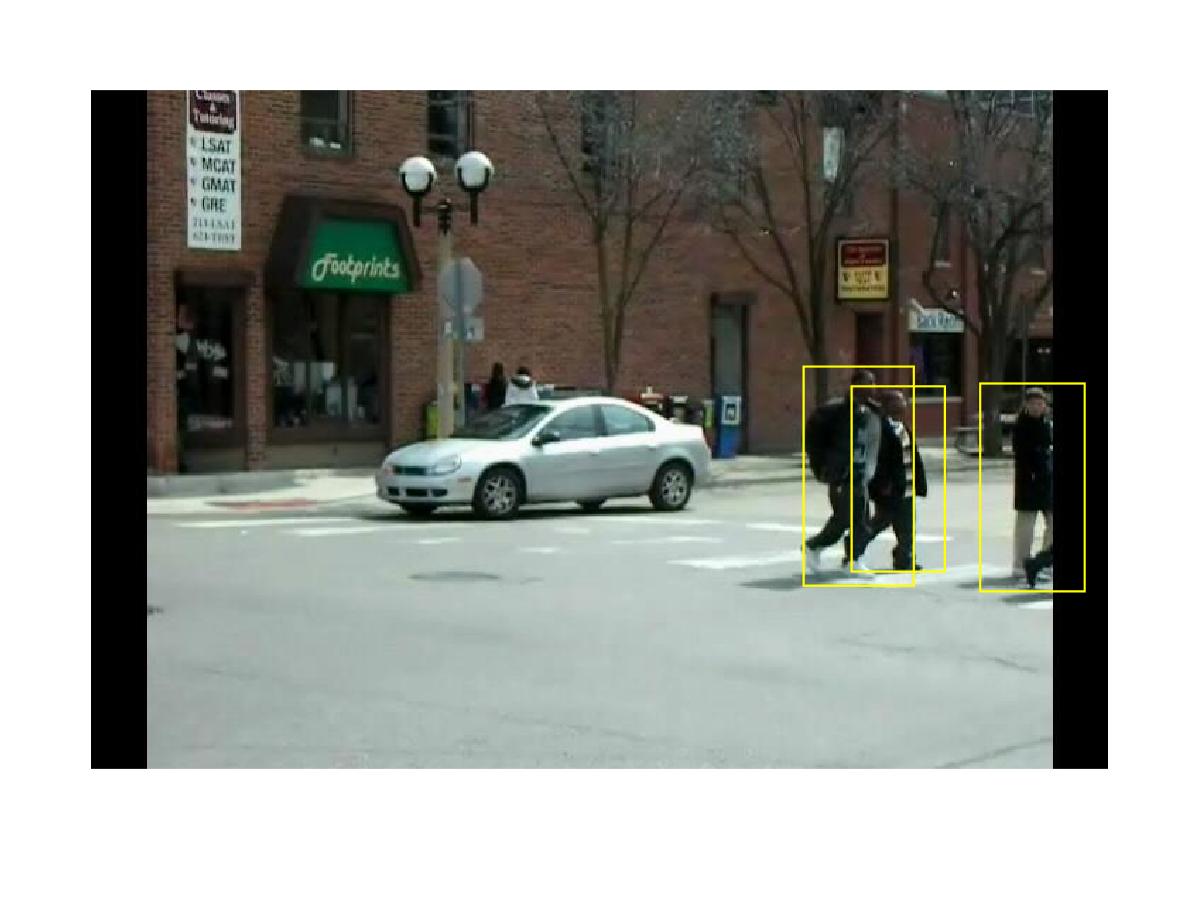}}
   \caption{Sample frames from 4 example sequences (in columns) from the collective activity dataset with the `Crossing scene level activity'. The colour of the bounding box indicates the activity class of each individual where yellow denotes `Crossing', green denotes `Waiting'  and blue denotes `Walking'. The sequences illustrate the challenges due to view point changes and visual similarity between the transition frames and the action frames (i.e 3rd column. transitions from `Crossing' to `Walking'). }
 \label{fig:vis_collective}
\end{figure}

In Figure \ref{fig:vis_collective} we visualise sample frames for 4 sequences from the collective activity dataset which contain the `Crossing' scene level activity. We highlight each pedestrian within a bounding box which is colour coded based on the individual activity performed where yellow denotes `Crossing',  green denotes `Waiting'  and blue denotes `Walking' activity classes. Note that the group activity label is assigned by considering the action that is performed by the majority of people in the sequence. These sequences clearly illustrate the challenges with the dataset. For the same scene level activity we observe significant view point changes. Furthermore there exists a high degree of visual similarity between the action transition frames and the action frames themselves. For example in 3rd column we observe such instances where the pedestrians transition from the `Crossing' to `Walking' classes. However, the proposed architecture has been able to overcome these challenges and generate accurate predictions. 

Comparing Table \ref{tab:table1} with Table \ref{tab:table2}, we observe a similar performance for \cite{ibrahim2016cvpr,CERN_cvpr} with the volleyball dataset due to the deficiencies in the two level modelling structure. 
In \cite{SRNN_wacv2018} and \cite{social_scene} the methods achieved improvements over \cite{ibrahim2016cvpr,CERN_cvpr} by pooling the hidden feature representation when predicting the group activities. However, these methods still utilise hand engineered loss functions for training the model. Our proposed GAN based model is capable of learning a mapping to an intermediate representation (i.e action codes) which is easily distinguishable for the activity classifier. The automatic loss function learning process embedded within the GAN objective synthesises this artificial mapping. Hence we are able to outperform the state-of-the-art methods in all considered metrics. 

With the results presented in Table \ref{tab:table2} we observe a clear improvement in performance over the baseline methods when considering players as 2 groups rather than 1 group. The 2 group representation first segments the players into the respective 2 teams using the ground truth annotations and then pools out the features from the 2 groups separately. Then these team level features are merged together for the group activity recognition. In contrast, the 1 group representation considers all players at once for feature extraction, rather than considering the two state approach. However this segmentation process is an additional overhead when these annotations are not readily available. In contrast the proposed MLS-GAN method receives all the player features together and automatically learns the contribution of each player for the group activity, outperforming both the 1 group and 2 group methods. We argue this is a result of the enhanced structure with the gated fusion units for the feature fusion process. Instead of learning a single static kernel for pooling out features from each player in the team, we attend over all the feature streams from both the player and scene levels, at that particular time step. This generates a system which efficiently varies the level of attention to each feature stream depending on the scene context.

\begin{figure}[htbp]
 \centering
    \subfigure[l-set]{\includegraphics[width=.3\linewidth,height=2.0cm]{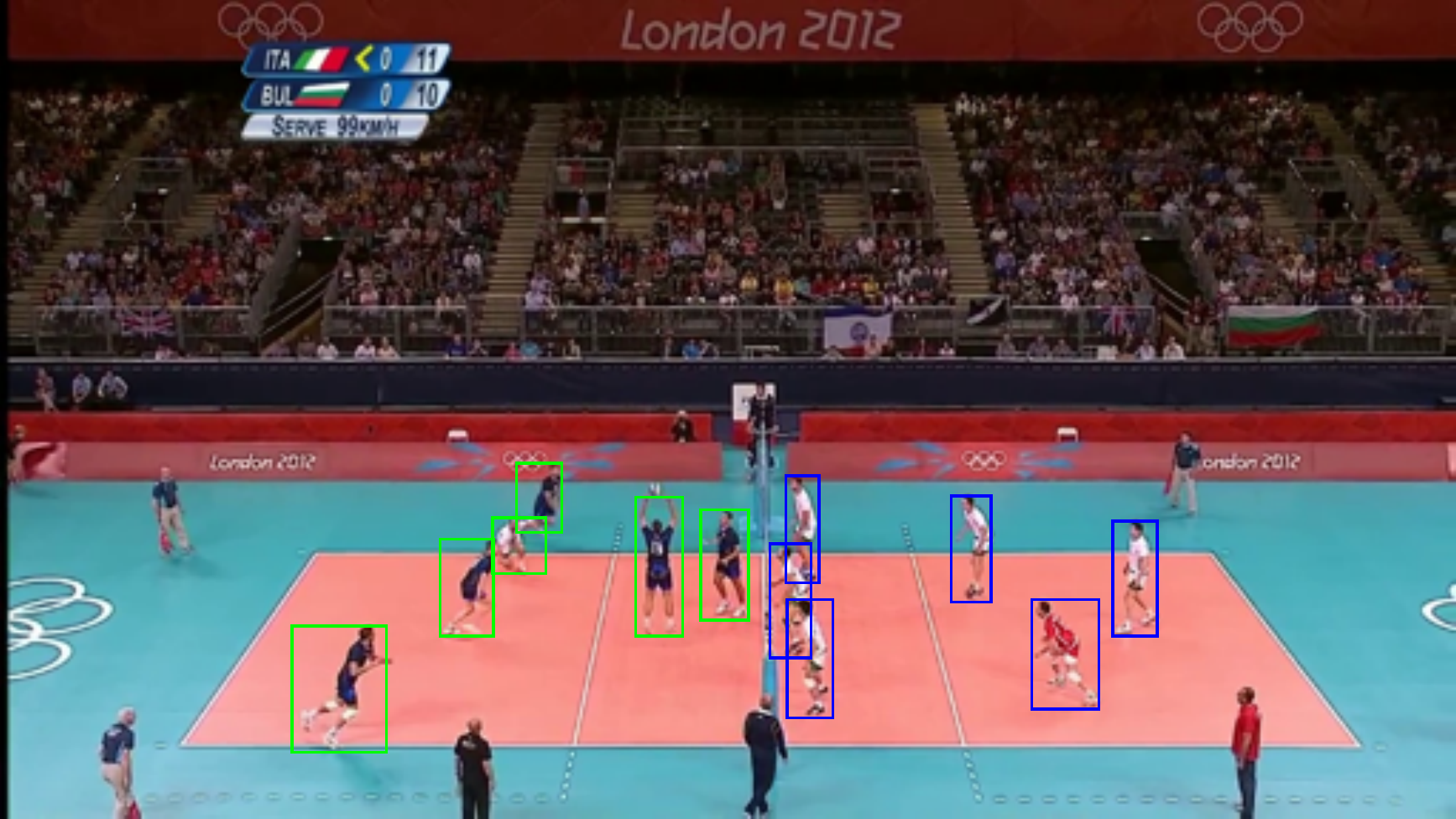}}
    \subfigure[l-pass]{\includegraphics[width=.3\linewidth,height=2.0cm]{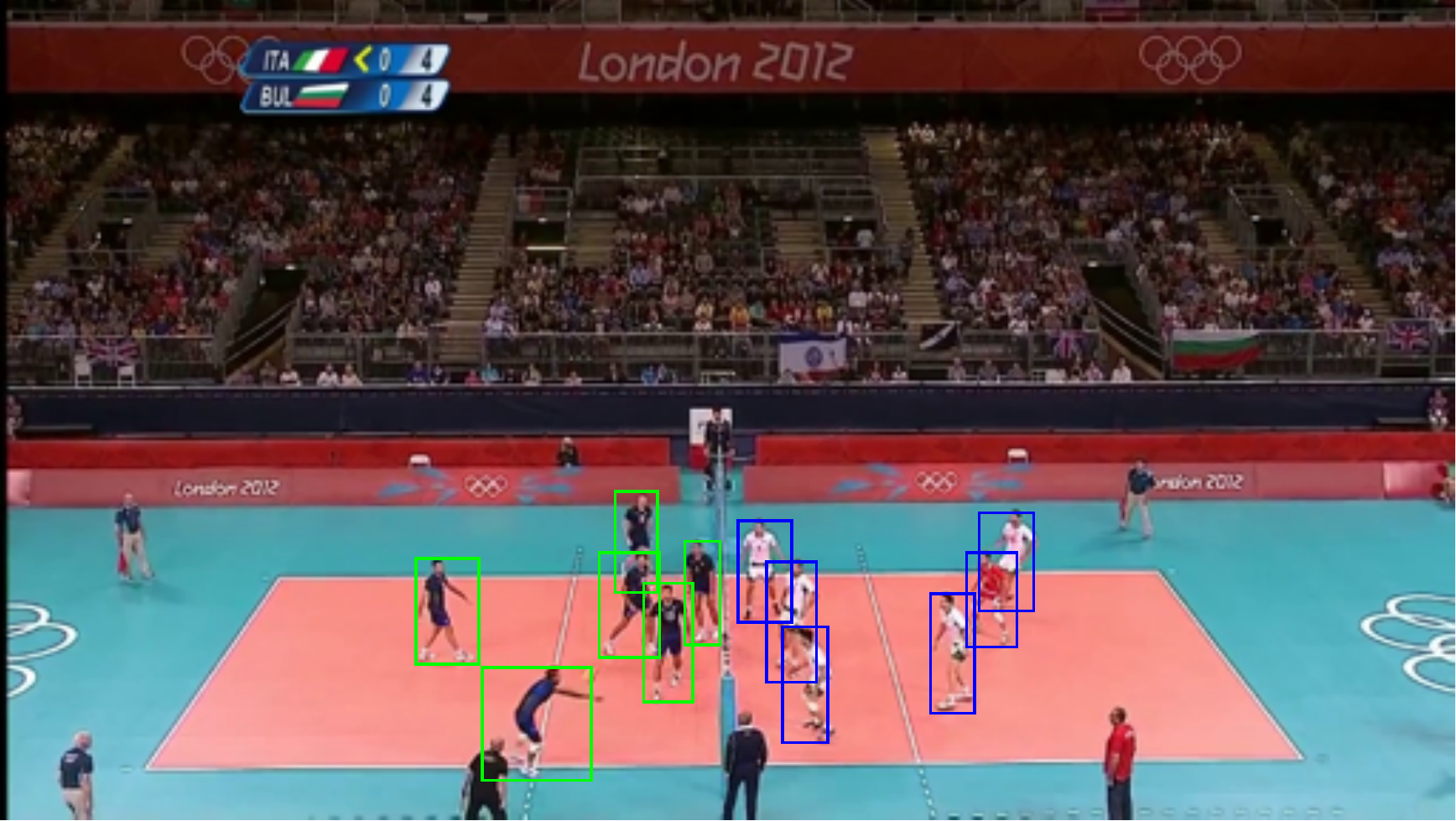}}
    \subfigure[l-spike]{\includegraphics[width=.3\linewidth,height=2.0cm]{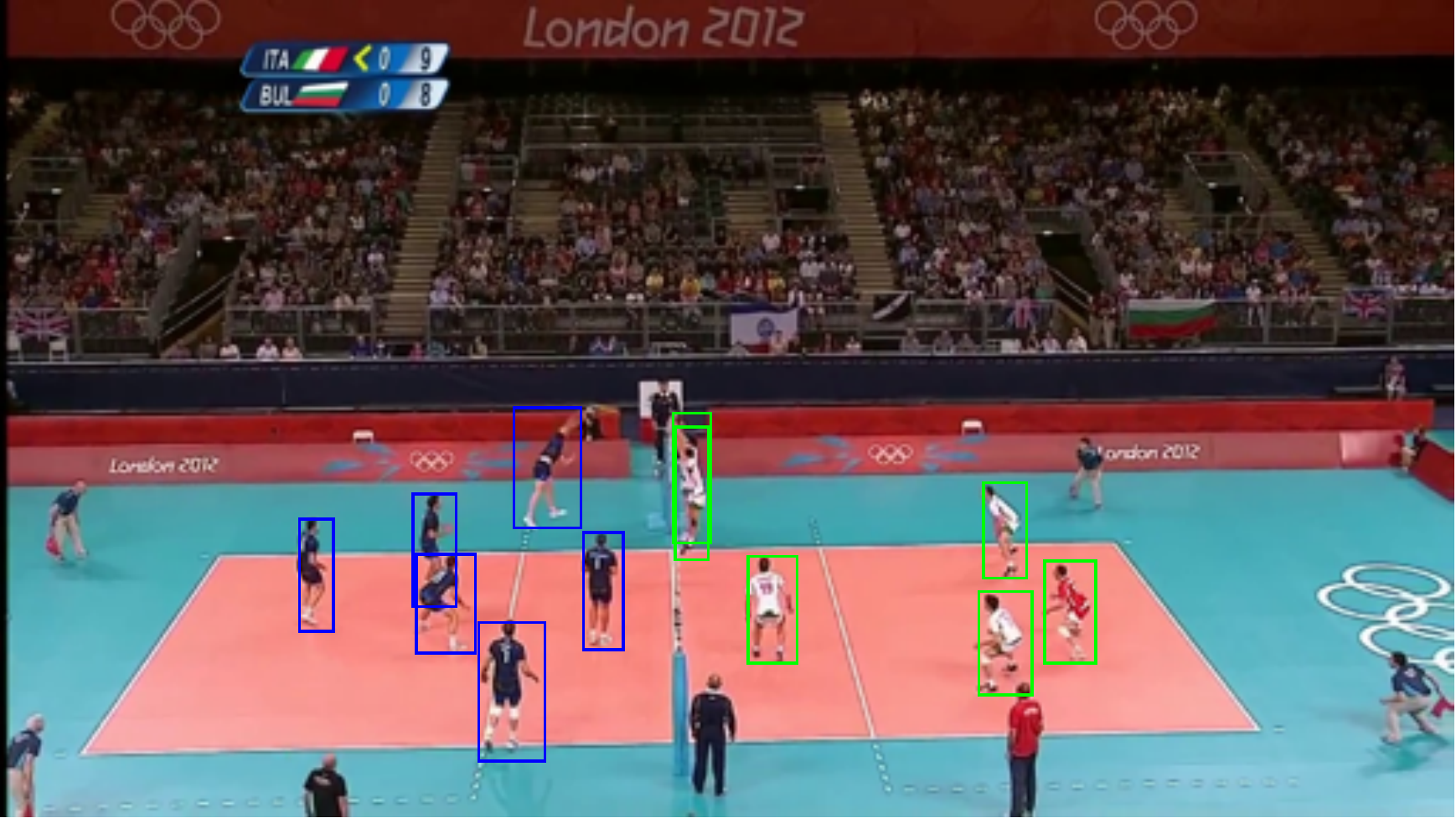}}
    \subfigure[r-set]{\includegraphics[width=.3\linewidth,height=2.0cm]{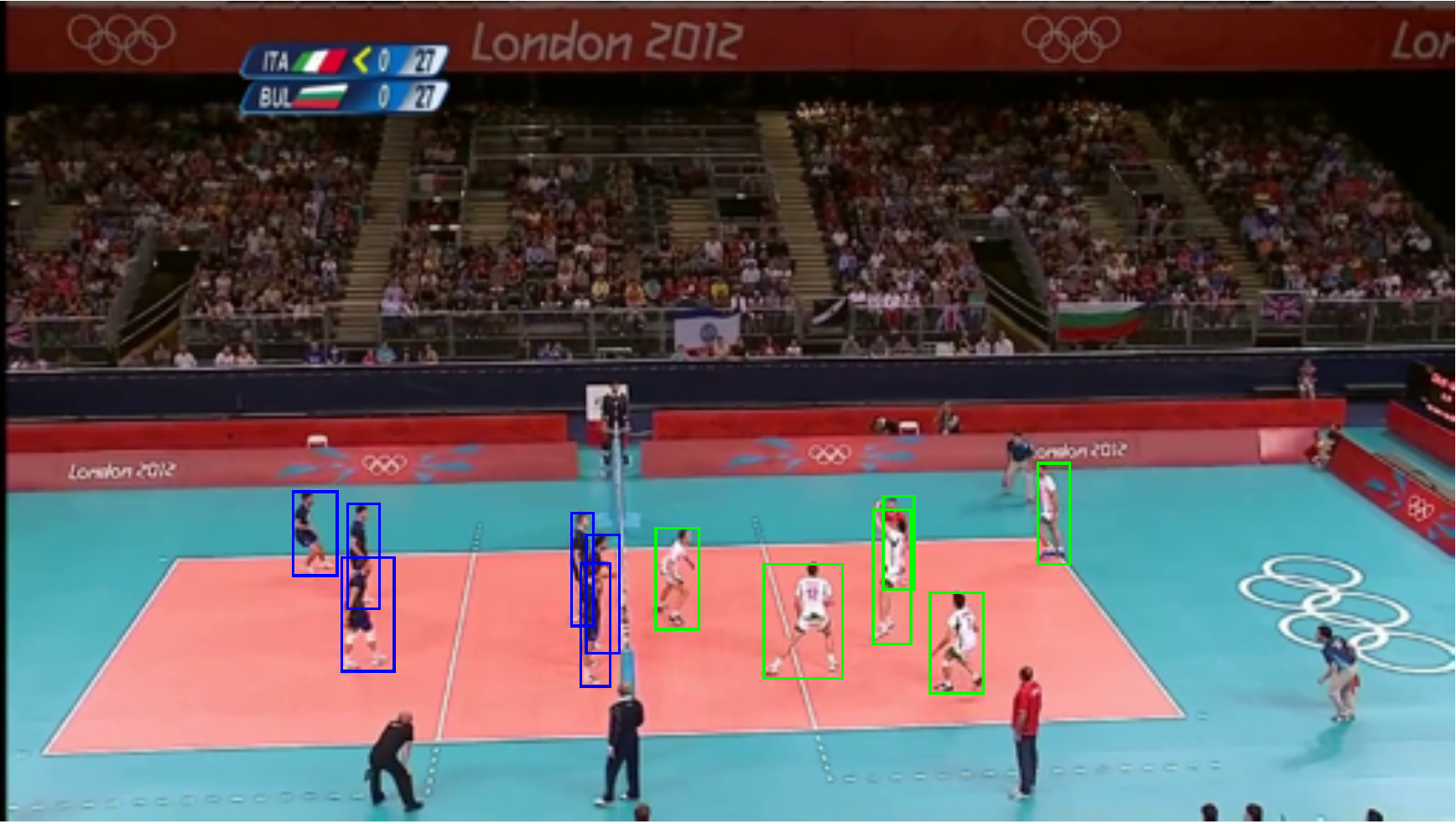}}
     \subfigure[r-spike]{\includegraphics[width=.3\linewidth,height=2.0cm]{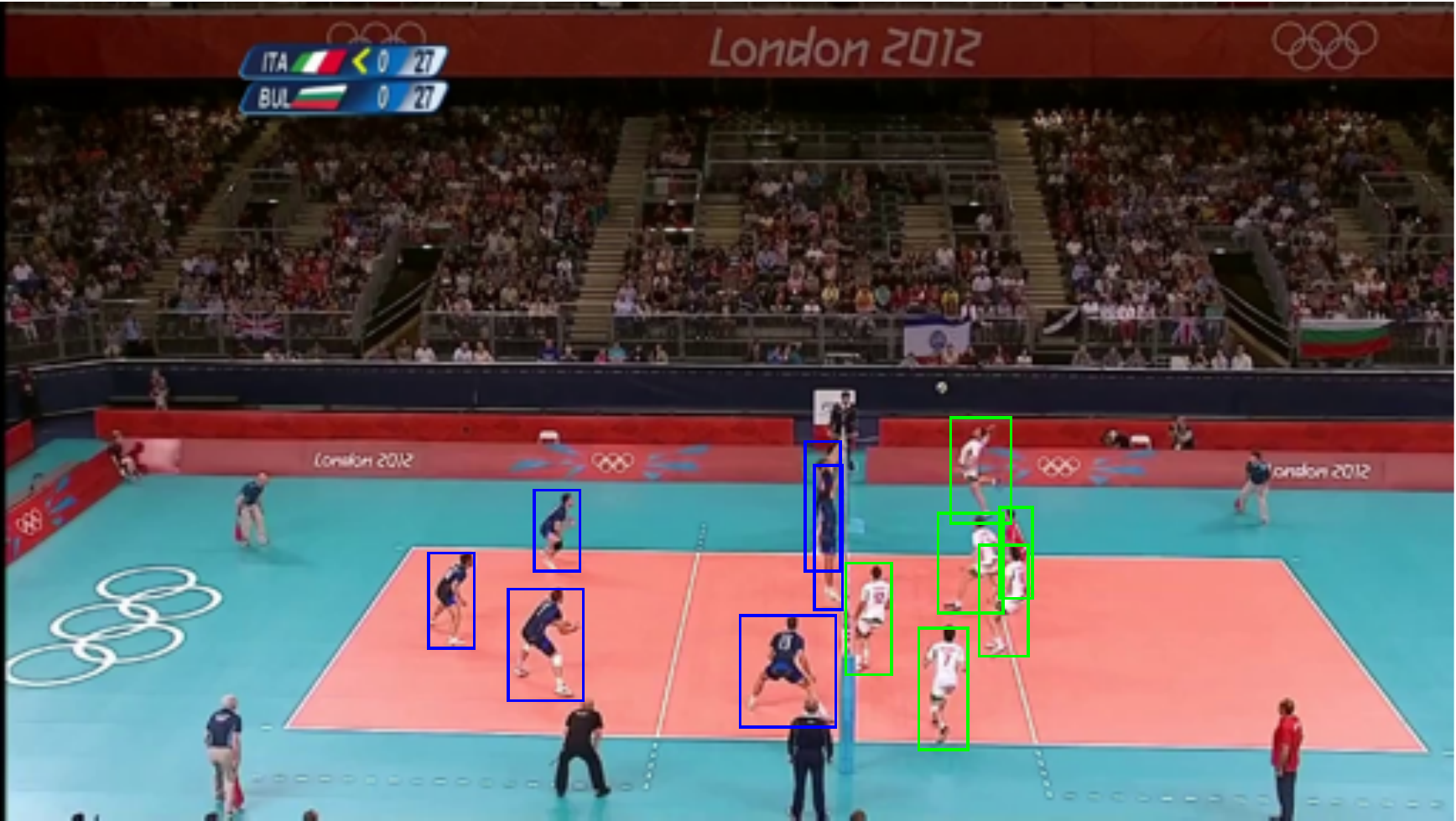}}
    \subfigure[r-pass]{\includegraphics[width=.3\linewidth,height=2.0cm]{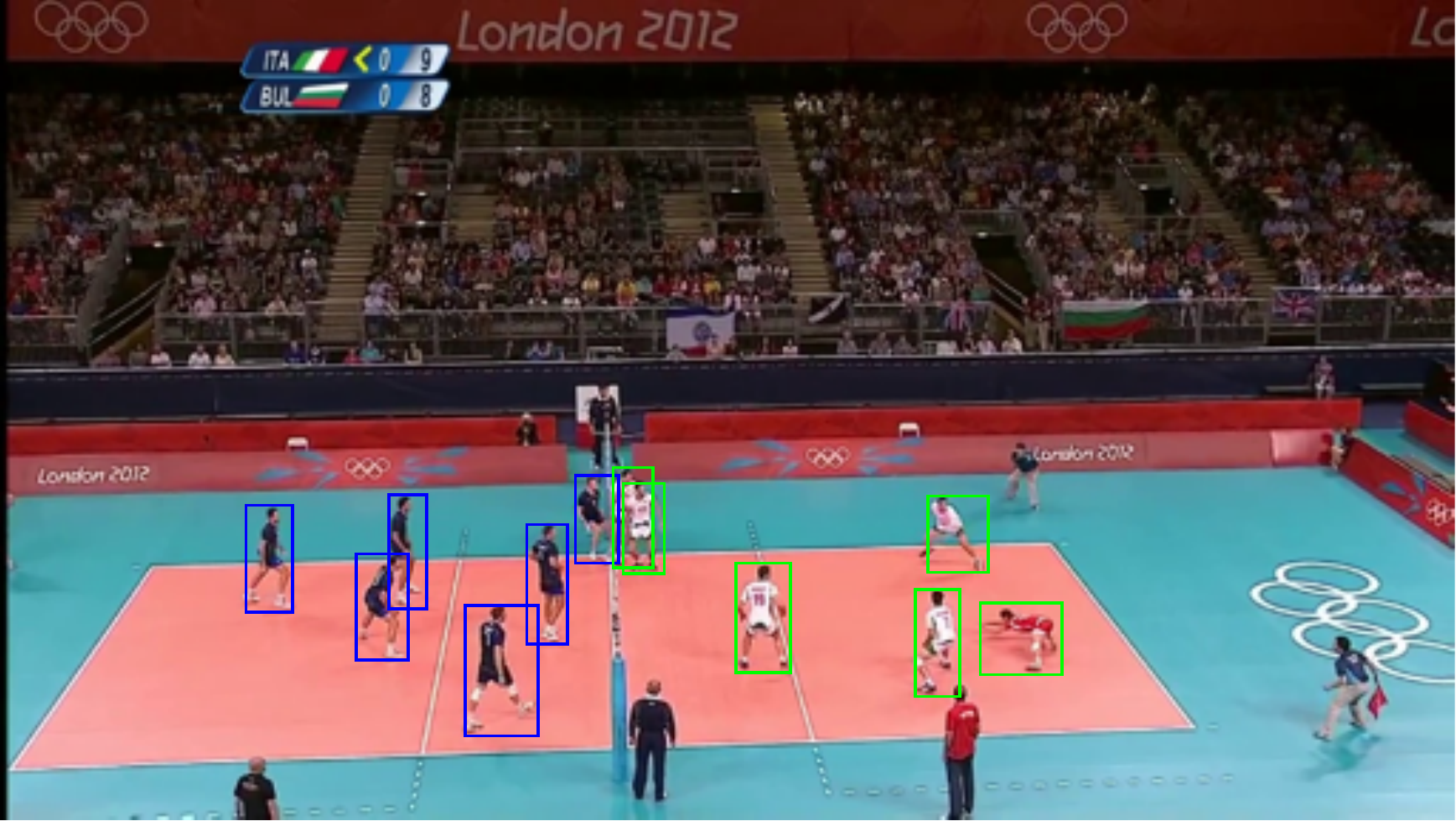}}
      \caption{Visualisations of the predicted group activities for the Volleyball dataset using the proposed MLS-GAN model.}
       \label{fig:vis_volleball}
\end{figure}

Figure \ref{fig:vis_volleball} visualises qualitative results from the proposed MLS-GAN model for the Volleyball dataset. Irrespective of the level of clutter and camera motion, the proposed model correctly recognises the group activity. 

\begin{table}[htbp]
\centering
\caption{Comparisons with the state-of-the-art for Volleyball Dataset \cite{ibrahim2016cvpr}. The first block of results (1 group) are for the methods considering all the players as a one group and the second block is for dividing players into two groups (i.e each team) first and extracting features from them separately. NA refers to unavailability of results.}
\label{tab:table2}

\begin{tabular}{|p{6cm}|c|c|}
\hline
                                                                                   Approach                              & MCA                               & MPCA                              \\ \hline \hline
                                                                                          2-layer LSTMs \cite{ibrahim2016cvpr}   (1 group)                      & 70.3                              & 65.9                              \\\hline
                                                                                           CERN \cite{CERN_cvpr}   (1 group)                                  & 73.5                              & 72.2                              \\ \hline
                                                                                           SRNN \cite{SRNN_wacv2018}    (1 group)                                 & 73.39                             & NA                                \\  \hline\hline

                                                                                          2-layer LSTMs \cite{ibrahim2016cvpr}  (2 group)                          & 81.9                              & 82.9                              \\ \hline
                                                                                          CERN \cite{CERN_cvpr}               (2 group)                      & 83.3                              & 83.6                              \\ \hline
                                                                                          SRNN \cite{SRNN_wacv2018}      (2 group)                               & 83.47                             &   NA                                \\\hline
                                                                                          Social Scene \cite{social_scene}    (2 group)                         & 89.90                              &    NA                               \\ \hline\hline
 \cellcolor[HTML]{C0C0C0}MLS-GAN & \cellcolor[HTML]{C0C0C0} \textbf{93.0}         & \cellcolor[HTML]{C0C0C0} \textbf{92.4}          \\ \hline
 \end{tabular}
\end{table}

\subsection{Ablation Experiments}

We further experiment with the collective activity dataset by conducting an ablation experiment using a series of a models constructed by removing certain components from the proposed \textit{MLS-GAN} model. Details of the ablation models are as follows:

\begin{enumerate}[label=\alph*]
\item \textit{G-GFU}: We use only the generator from \textit{MLS-GAN} and trained it to predict group activity classes by adding a final softmax layer. This model learns through supervised learning using categorical cross-entropy loss.  Further, we removed the Gated Fusion Unit (GFU) defined in Eq. \ref{eq:6} to Eq. \ref{eq:9}. Therefore this model simply concatenates the outputs from each stream. 
\item  \textit{G}: The generator model plus the GFU trained in a fully supervised model as per the \textit{G-GFU} model above. 
\item  \textit{cGAN-(GFU and $\hat z$)}: a conditional GAN architecture where the generator model utilises only the person-level features (no scene-level features), and does not utilise the GFU mechanism for feature fusion. However the discriminator model $D$ still receives the scene level image and the generated action code as the inputs.
\item  \textit{cGAN-GFU}: a conditional GAN architecture which is similar to the proposed \textit{MLS-GAN model}, however does not utilise the GFU mechanism for feature fusion. 
\item  \textit{MLS-GAN- $\hat z$}: \textit{MLS-GAN} architecture where the generator utilises only the person-level features for action code generation. The discriminator model is as in \textit{cGAN-(GFU and $\hat z$)}. As per \textit{cGAN-(GFU and  $\hat z$)}, the discriminator still recieves the scene level image.

\end{enumerate}

\begin{table}[htbp]
\centering
\caption{Ablation experiment results on Collective Activity dataset \cite{choi2009}.}
\label{tab:table3}
\begin{tabular}{|p{4.5cm}|c|c|}
\hline
                                                                                   Approach                              & MCA                               & MPCA                              \\ \hline \hline

                                                                                          \textit{G-GFU}                           &           58.9                    &       58.7                        \\ \hline
                                                                                          \textit{G}                      &               61.3                &               60.5                    \\ \hline \hline
                                                                                          \textit{cGAN-(GFU and $\hat z$)}                   &       88.4                        &          87.7                     \\  \hline 
                                                                                          \textit{cGAN-GFU}                   &            89.5                   &        88.3                       \\  \hline 
                                                                                          \textit{MLS-GAN- $\hat z$}                   &       91.2                        &          90.8                     \\  \hline \hline
                                                                                         
\cellcolor[HTML]{C0C0C0}MLS-GAN & \cellcolor[HTML]{C0C0C0}\textbf{91.7} & \cellcolor[HTML]{C0C0C0}\textbf{91.2} \\ \hline
\end{tabular}
\end{table}

When analysing the results presented in Table \ref{tab:table3} we observe significantly lower accuracies for methods \textit{G-GFU} and \textit{G}. Even though a slight improvement of performance is observed with the introduction of the GFU fusion strategy, still we observe a significant reduction in performance. We believe this is due to the deficiencies with the supervised learning process where we directly map the dense visual features to a sparse categorical vector. However, with the other variants and the proposed approach we learn an objective which maps the input to an intermediate representation (i.e action codes) which is easily distinguishable by the classifier. The merit of the intermediate representation is shown by the performance gap between \textit{G} and the \textit{cGAN-(GFU and $\hat z$)}, which we further enhance in \textit{cGAN-GFU} by including scene information alongside the features extracted for the individual agents. This allows the GAN to understand the spatial arrangements of the actors when determining the group activity. Comparing \textit{cGAN-GFU} and \textit{MLS-GAN- $\hat z$}, we can also see the value of the GFU which is able to better combine data from the individual agents. Finally by utilising both person-level and scene-level features and combining those through proposed GFUs the proposed \textit{MLS-GAN} model attains better recognition results.

We would like to further compare the performance of non-GAN based models \textit{G-GFU} and \textit{G} with the results for the deep architectures in Table \ref{tab:table1}. Methods such as the 2-layer LSTMs \cite{ibrahim2016cvpr} and CERN \cite{CERN_cvpr} have been able to attain improved performance compared to \textit{G-GFU} and \textit{G}, however with the added expense of the need for hand annotated individual actions in the database. In contrast, with the improved GAN learning procedure the same architectures (i.e \textit{cGAN-(GFU and $\hat z$)}, \textit{cGAN-GFU} and \textit{MLS-GAN- $\hat z$}) have been able to achieve much better performance without using those individual level annotations. 

In order to further demonstrate the discriminative power of the generated action codes we directly classified the action codes generated by \textit{cGAN-(GFU and $\hat z$)} model. We added a softmax layer to the generated model and tried directly classifying the action codes. We trained only this added layer by freezing the rest of the network weights. We obtained 90.7 MPCA for the collective activity dataset. Comparing this with the ablation model $G$ in Table \ref{tab:table3} (the generator without the GAN objective, trained using only the classification objective), the reported MPCA value is 60.5. Hence it is clear that the additional GAN objective makes a substantial contribution.

\subsection{Time Efficiency}
We tested the computational requirements of the MLS-GAN method using the test set of the Volleyball dataset \cite{ibrahim2016cvpr} where the total number of persons, $N$, is set to 12 and each sequence contains 10 time steps. Model generates 100 predictions in 20.4 seconds using a single core of an Intel E5-2680 2.50 GHz CPU.

\section{Conclusions}

In this paper we propose a Multi-Level Sequential Generative Adversarial Network (\textit{MLS-GAN}) which is composed of LSTM networks for capturing separate individual actions followed by a gated fusion unit to perform feature integration, considering long-term feature dependancies. We allow the network to learn both person-level and scene-level features to avoid information loss on related objects, backgrounds, and the locations of the individuals within the scene. With the inherited ability to learn both features and the loss function automatically, we employ a semi supervised GAN architecture to learn an intermediate representation of the scene and person-level features of the given scene, rendering an easily distinguishable vector representation, an action code, to represent the group activity. Our evaluations on two diverse datasets, Volleyball and Collective Activity datasets, demonstrates the augmented learning capacity and the flexibility of the proposed \textit{MLS-GAN} approach. Furthermore, with the extensive evaluations it is evident that the combination of scene-level features with person-level features is able to enhance performance by a considerable margin.

\end{document}